\pdfoutput=1
\documentclass{article}

    \PassOptionsToPackage{numbers, compress}{natbib}

\usepackage[final]{neurips_2023}

\usepackage{geometry}
\usepackage{longtable}
\usepackage{array}

\usepackage[utf8]{inputenc} 
\usepackage[T1]{fontenc}    
\usepackage{hyperref}       
\usepackage{url}            
\usepackage{booktabs}       
\usepackage{amsfonts}       
\usepackage{nicefrac}       
\usepackage{microtype}      
\usepackage{xcolor}         
\usepackage{soul}
\usepackage{fancyhdr}

\usepackage[pdftex]{graphicx}
\graphicspath{ {figures/} }
\usepackage{wrapfig}
\usepackage{caption}
\usepackage{subcaption}
\usepackage{todonotes}
\usepackage{xcolor}
\usepackage[normalem]{ulem}
\usepackage{enumitem}
\usepackage{subcaption}

\usepackage{titlesec}
\titlespacing*{\section}{0pt}{0.1\baselineskip}{0.2\baselineskip}
\titlespacing*{\subsection}{0pt}{0.1\baselineskip}{0.2\baselineskip}
\titlespacing*{\subsubsection}{0pt}{0.1\baselineskip}{0.2\baselineskip}

\title{Reconstructing Materials Tetrahedron: Challenges in Materials Information Extraction}

\author{
    Kausik Hira$^{1}$, Mohd Zaki$^{2}$, Dhruvil Sheth$^{1}$, \\
    \hspace*{0.5cm} \textbf{Mausam}$^{1}$, \textbf{N. M. Anoop Krishnan}$^{1,2}$\\
    $^{1}$Yardi School of Artificial Intelligence, $^{2}$Department of Civil Engineering \\
    Indian Institute of Technology Delhi \\
    \small{\texttt{\{kausikhira, mohdzaki1995, dhruvilsheth01\}@gmail.com}} \\
    \small{Corresponding authors: \texttt{\{mausam, krishnan\}@iitd.ac.in}} \\
}

\begin{document}

\maketitle
\begin{abstract}
The discovery of new materials has a documented history of propelling human progress for centuries and more. The behaviour of a material is a function of its composition, structure, and properties, which further depend on its processing and testing conditions. Recent developments in deep learning and natural language processing have enabled information extraction at scale from published literature such as peer-reviewed publications, books, and patents. However, this information is spread in multiple formats, such as tables, text, and images, and with little or no uniformity in reporting style giving rise to several machine learning challenges. Here, we discuss, quantify, and document these challenges in automated information extraction (IE) from materials science literature towards the creation of a large materials science knowledge base. Specifically, we focus on IE from text and tables and outline several challenges with examples. We hope the present work inspires researchers to address the challenges in a coherent fashion, providing a fillip to IE towards developing a materials knowledge base.
\end{abstract}

\section{Introduction}
\label{sec:introduction}


Understanding a material's behavior requires knowledge about its composition, properties, processing and testing protocols, and microstructure---represented as the materials science (MatSci) tetrahedron (see Fig. \ref{fig:graph_abs}). These different aspects of a material are reported by researchers in peer-reviewed publications, patents, and other scientific documents. Recently, there have been several attempts to exploit the advances in machine learning (ML) and artificial intelligence (AI) towards automated information extraction (IE) from literature\cite{elsa_review_nlp_matsci_ie,glassomics,ravinder2021artificial,mascqa-2024}. These include the development of materials specific language models~\cite{gupta2022matscibert,huang2022batterybert,trewartha2022quantifying, gupta-etal-2023-discomat}, rule-based systems~\cite{swain2016chemdataextractor,mavracic2021chemdataextractor,zaki2022extracting,venugopal2021looking,zaki2022natural}, IE from tables~\cite{gupta-etal-2023-discomat,zhao2022database,zhao2023opticalbert}, and IE from images~\cite{mukaddem2019imagedataextractor,zhang2022image,zaki2023cementron,zaki2022interpretable}. The widely varying information expression styles in research papers makes the automated MatSci IE a challenging task. Most of the works have focused on IE in a specific domain; hence, the transferability to different materials is not explored. Moreover, no consolidated work exists that explores the specific challenges associated with IE in MatSci and the gain associated with solving these challenges, which provides a clear direction to the researchers regarding the areas that require increased attention.


 We thoroughly review MatSci articles to identify IE challenges towards completing the materials tetrahedron (see Fig.~\ref{fig:graph_abs}). We also highlight some of the major challenges toward the development of a ``universal'' MatSci knowledge base linking the extracted information from multiple sources and forms of data---structured, semi-structured, and unstructured. Indeed, millions of scientific documents exist reporting information about various materials known to humans. Thus, the automated development of MatSci IE will lead to a rich knowledge base on materials. The outline of the paper is as follows: First, we explain the methodology of collecting papers for review and annotation process. Then, in the results and discussion sections, we investigate the proportion of each of the entities, such as composition, structure, properties, processing, and testing conditions, reported in tables or text of the articles, followed by the challenges faced in their extraction.  We quantify how frequently a challenge occurs to motivate researchers to gauge the amount of information that will be obtained after solving the respective challenges. We further identify the challenges in extracting and connecting the information from text and tables and among different tables belonging to the same MatSci research papers. Note that the challenges reported for extracting compositions from tables are verified by the present IE models, and only those that are unaddressed or solved unsatisfactorily are reported in the main text, whereas some of the existing challenges that have been resolved satisfactorily are documented in the appendix. In our study, DiSCoMaT~\cite{gupta-etal-2023-discomat} was employed as the IE model for extracting compositions from tables, recognized as the most effective IE model for this purpose~\cite{mascqa-2024}. Concurrently, GPT-4 was utilized to extract compositions from textual content in our study. For extracting properties from MatSci tables, we could not find any domain-specific IE model, but we believe that the challenges reported are valid for any IE models. We have also provided reasons and examples to elaborate on the same. Regarding IE from text to complete materials tetrahedron, we have highlighted examples where existing IE models also make mistakes. Finally, we provide some guidelines for presenting machine and human-friendly tables that enable automated MatSci IE from research papers.

\section{Methodology}
\label{sec:methods}

\begin{figure}
    \centering
    \includegraphics[width=1.0\textwidth, page=2]{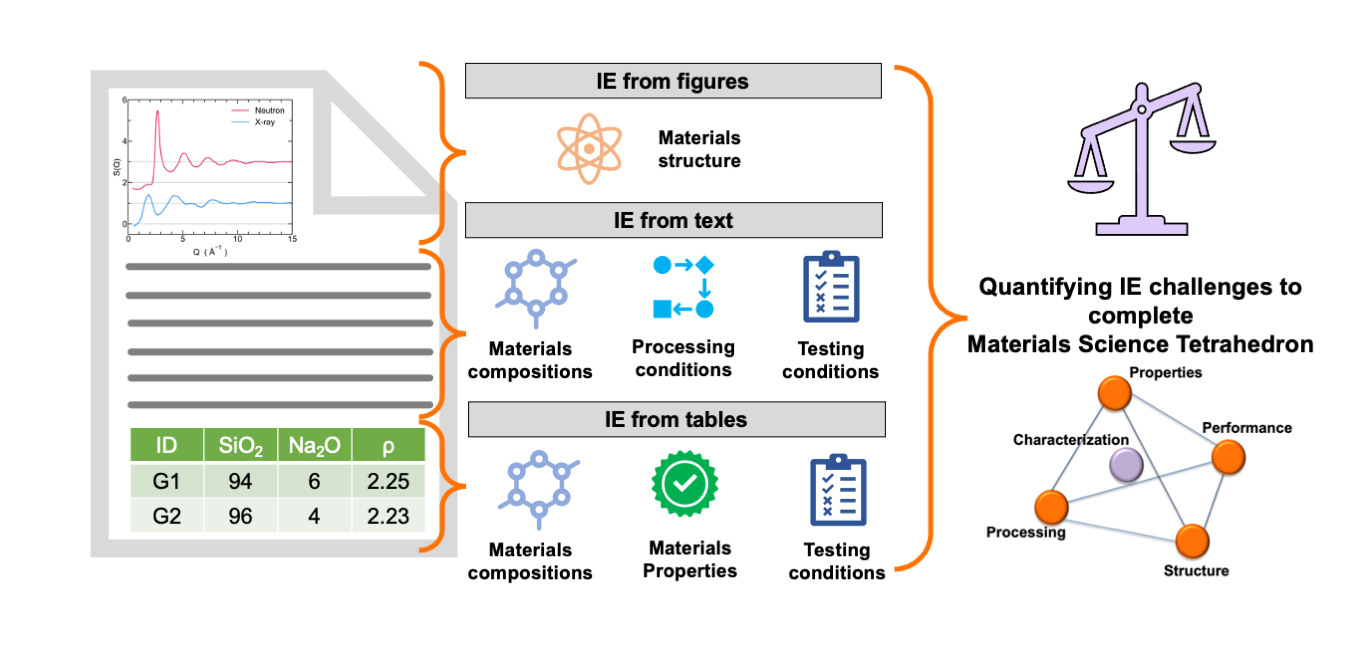}
    \caption{Quantifying challenges in information extraction from different elements of a research paper such as text, tables, and figures.}
    \label{fig:graph_abs}
    \vspace{-0.2in}
\end{figure}

To elucidate the challenges, we referred to a dataset of 2536 peer-reviewed publications on MatSci. This dataset is taken from recent work on IE from tables~\cite{gupta-etal-2023-discomat}, where the authors used distant supervision to annotate tables from research papers based on respective compositions present in INTERGLAD~\cite{interglad}. The tables in val and test data were annotated manually by indicating the relevant rows and columns that should be used to extract material compositions. Fig. ~\ref{fig:graph_abs} shows different sections of the paper where these different components are majorly reported. The statistics of each challenge were computed by randomly taking 50/100 tables from the manually annotated val and test dataset. In cases where this was not applicable, we further performed manual annotation on an additional 50 papers or 100 relevant tables selected randomly from the corpus. For instance, we randomly selected 100 composition tables from the manual annotation in the existing dataset for composition extraction. However, no such manual annotation was available for properties. For this problem, we selected 100 random property tables from the corpus and manually annotated the frequency of the challenges in property extraction. Note that all the challenges and their reported frequencies are based on manual annotation, which is more reliable than any ML-based technique, such as distant supervision. Further, we manually analyzed tables or text for the occurrence of each of the entities, such as composition, structure, and property. All the results and data associated with the annotation process are shared in the following \href{https://github.com/M3RG-IITD/MatSci-IE-Challanges}{link}.

\section{Results and Discussions}
\label{sec:res_n_dis}
    
    
    


\begin{figure}
    \centering
    \includegraphics[width=0.50\textwidth,page=2]{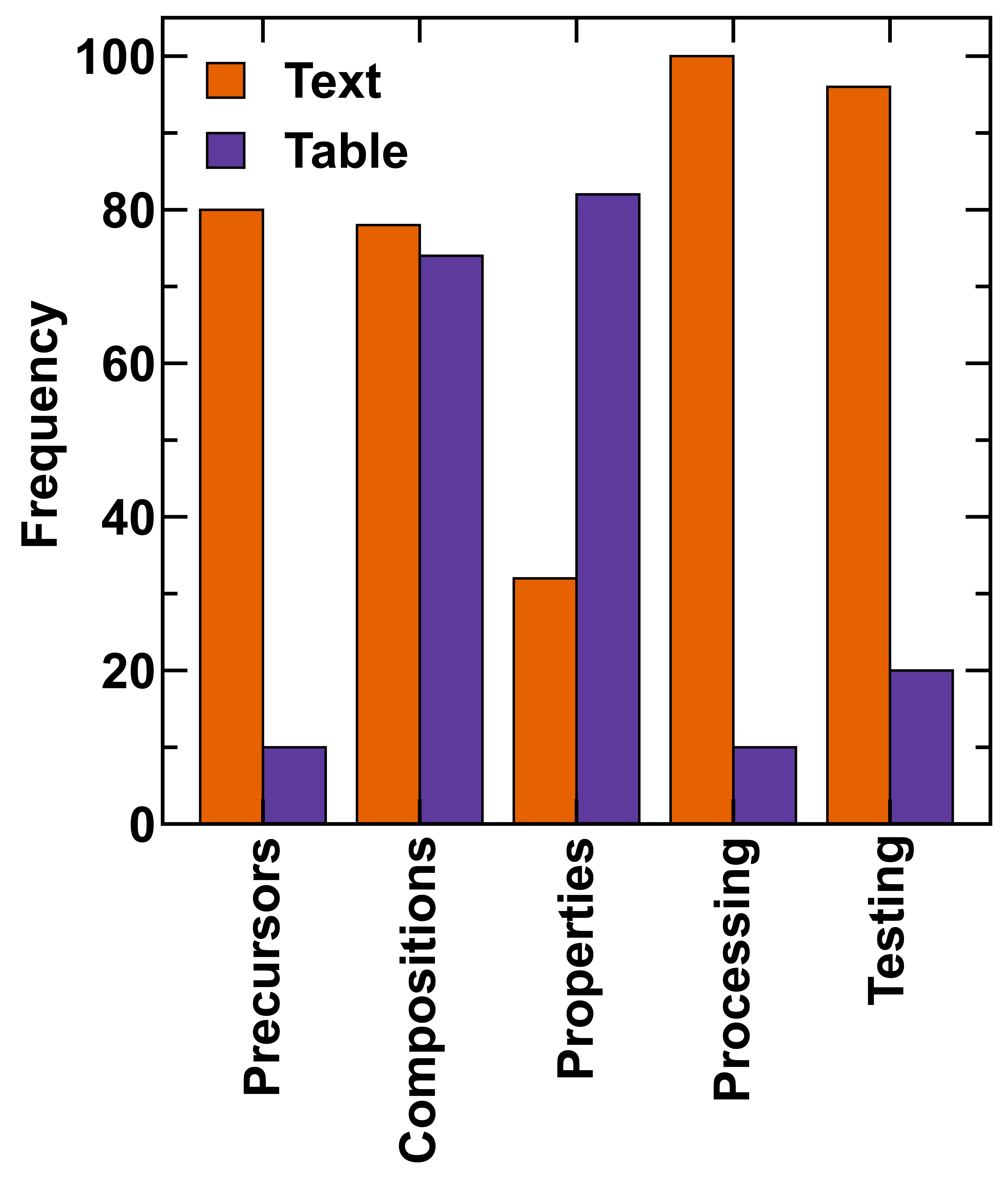}
    \caption{Occurrence of information regarding precursors(raw materials), compositions, properties, processing, and testing conditions in MatSci papers.}
    \label{fig:info_location}
    \vspace{-0.1in}
\end{figure}

Figure~\ref{fig:info_location} shows the percentage of papers reporting raw materials (precursors), compositions, properties, processing, and testing methods in text and tables. Note that the same information could be reported in both text and table and hence, the percentages may add to more than 100. Although 78\% and 74\% of papers had compositions in text and tables, respectively, an in-depth analysis revealed that only 33.21\% of the total compositions were reported in the text, whereas 85.92\% of compositions were present in tables. The overlap exists due to the same composition being mentioned in both text and tables. 82\% articles report properties in tables (see Fig.~\ref{fig:info_location}). Processing and testing conditions are mostly reported in the text, while in 80\% articles, precursors are mentioned in the text. In the following sections, we discuss these aspects in detail.

\subsection{Composition extraction}
\vspace{-0.1in}
    
    
    




Since the majority of the material compositions are reported in tables, we first discuss the challenges in extracting compositions from tables. This is followed by the discussion on IE from text.
\subsubsection{Extracting compositions from tables} 
\label{comp_table_extraction}
\vspace{-0.1in}
Here, we summarize the major challenges in composition extraction from tables. To this extent, we investigated 100 randomly selected composition tables from the manually annotated data to report the frequency of occurrence of each challenge.

\begin{figure}
    \centering
    \includegraphics[width=0.9\textwidth,page=3]{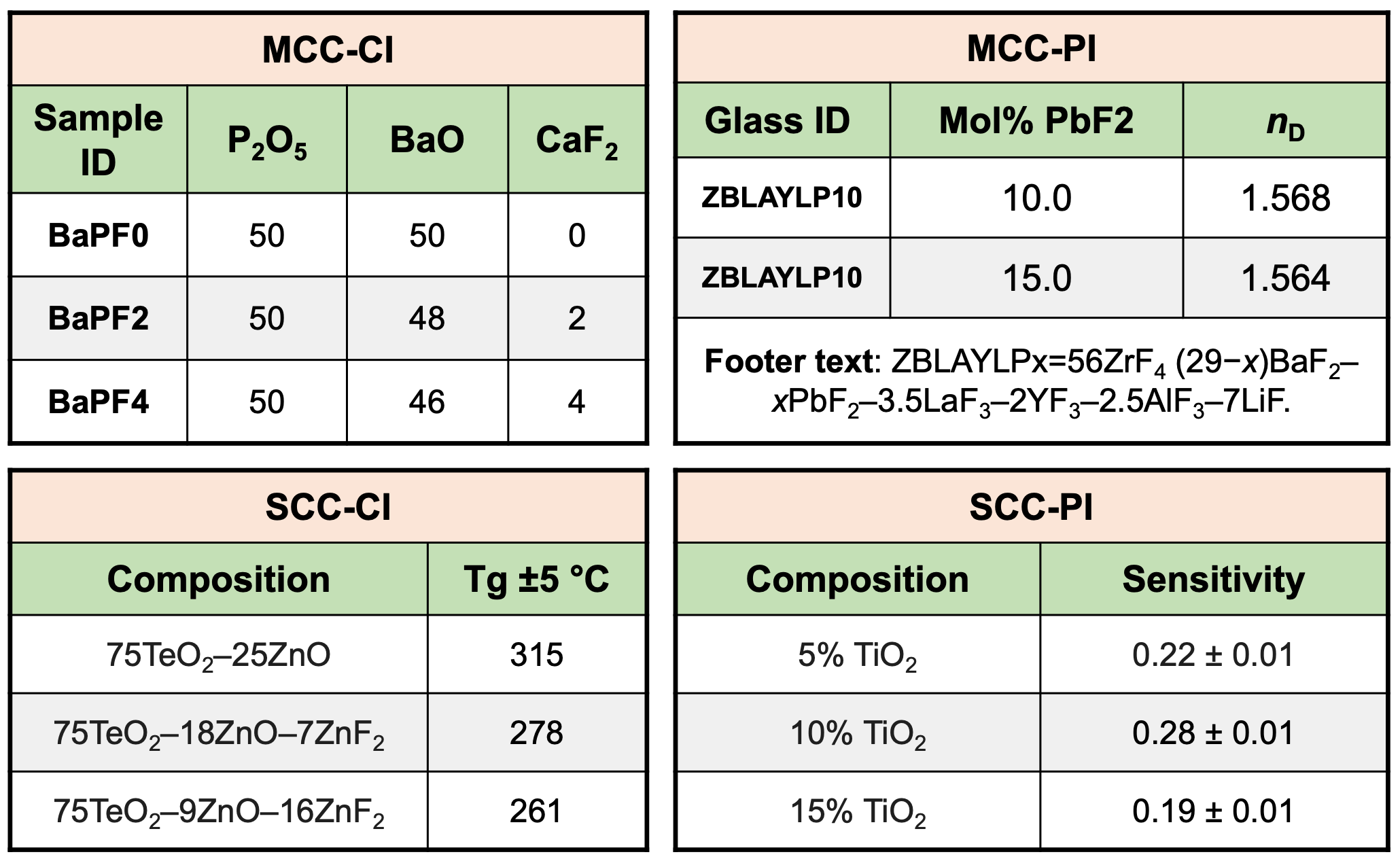}
    \caption{Classification of composition tables in single-cell composition (SCC) and multi-cell composition (MCC) with complete information (CI) and partial information (PI).}
    \label{fig:class_comp_tab}
\end{figure}

\textbf{a. Variation in table structure and information content:} An analysis of 100 random MatSci composition tables revealed that these tables do not follow any standard structure. Accordingly, following earlier schema proposed by Gupta et al.~\cite{gupta-etal-2023-discomat}, composition tables can categorized into two broad categories---multi-cell composition (MCC) and single-cell composition (SCC). These are further subdivided into tables containing complete information (CI) and partial information (PI). When the entire composition is written inside a single cell, it is classified as an SCC table, whereas when the composition is written across multiple cells of the table by reporting the value of each constituent (compounds or elements) of the composition in separate cells, it is defined to be an MCC table. If the table contains all the information regarding the constituents of the material, they are classified as CI tables (complete information). Alternatively, if only some of the constituents are mentioned in the table for the material, they are PI tables. In the latter case, we need to extend the analysis to the text of the article to extract the full composition. Fig. \ref{fig:class_comp_tab} illustrates all 4 types of tables~\cite{nazabal2003oxyfluoride,zaharescu2008tio2,narayanan2015thermal, jestin2003viscosity}. The most prevalent composition table types are MCC-CI (36\%), followed by SCC-CI (30\%). PI tables are less common, with 24\% being MCC-PI and the remaining 10\% being SCC-PI. Note that this distribution may also vary significantly depending on the material types. For instance, it is common practice in alloys to skip the major element while describing the composition in a table. In previous work by Gupta et al.~\cite{gupta-etal-2023-discomat}, while an F1 score of 78.21\% and 65.41\% have been achieve for extraction from SCC-CI and MCC-CI tables, respectively, an F1 score of only 51.66\% has been achieved for extraction from MCC-PI. Although the researchers have not explicitly focused on SCC-PI, we used their best model for SCC-PI tables and obtained 47.19\% as the F1 score. Hence, there is a significant scope for improvement in extracting compositions from PI tables.

\begin{figure}
    \centering
    \includegraphics[width=0.9\textwidth,page=4]{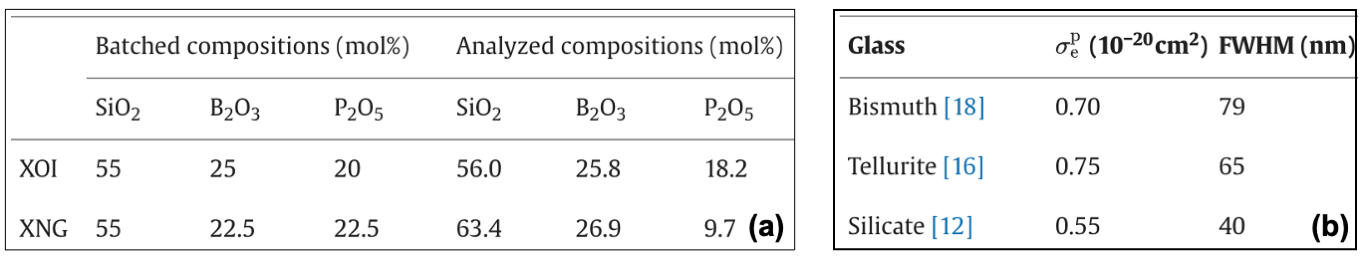}
    \caption{Example of tables: (a) mentioning nominal (batch) and analyzed composition, (b) having references to other papers}
    \label{fig:nom_ana}
\end{figure}

\textbf{b. Presence of nominal and experimental compositions}: While the nominal composition is the amount of chemicals taken initially to prepare the material, analyzed/experimental composition refers to the actual composition of material obtained after analyzing the manufactured material (see Fig. \ref{fig:nom_ana}(a))~\cite{YOUNGMAN2000111,PETKOV1999150}. Our analysis revealed that in 3\% of the tables, both nominal and analyzed/experimental compositions are reported. These values are not reported in any fixed pattern, making it difficult to correctly separate the nominal and analyzed compositions after extraction.

\begin{figure}
    \centering
    \includegraphics[width=0.9\textwidth]{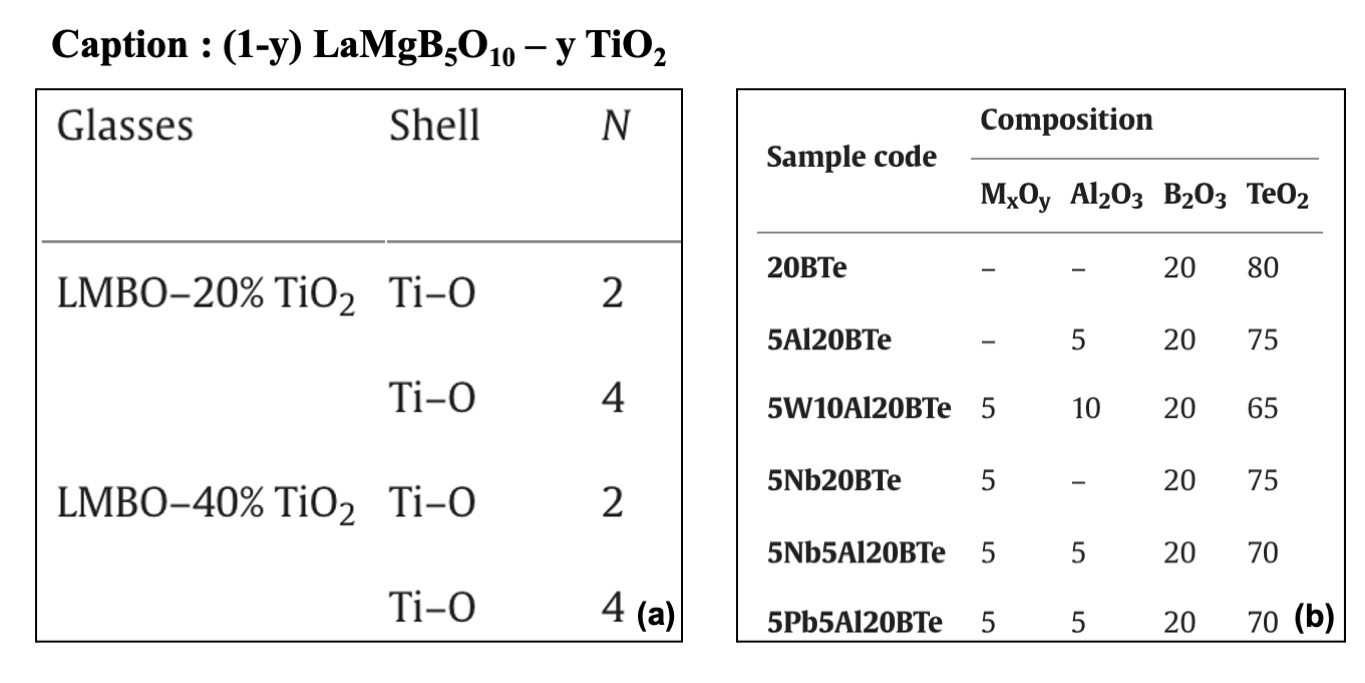}
    \caption{(a) Table with composition mentioned as acronyms in ID (first column). (b) The value of variable `M' needs to be inferred from the material IDs.}
    \label{fig:id2comp}
\end{figure}

\textbf{c. Compositions and related info inferred from other documents:} In some tables, the details of the glasses studied are not explicitly mentioned; rather, references to previous research publications which use the same material are provided in the tables or their captions (see Fig. \ref{fig:nom_ana}(b)). Thus, the composition or the other associated information of the material which is missing in the current publication must be extracted from the cited work, which then must be combined with the relevant information of the material present in the current work. We found references about different entities of the material in 11 tables~\cite{BRASIL20047,CHEN200312}. 4 out of the 11 tables have not explicitly mentioned compositions, due to which the IE model~\cite{gupta-etal-2023-discomat} was unsuccessful in obtaining the desired compositions. 


\textbf{d. Composition inferred from material ID:} We observed that 10\% of the total composition tables contain IDs with essential material composition information. In 60\% of these tables, DiSCoMaT~\cite{gupta-etal-2023-discomat} failed to extract the compositions correctly. Most of these tables did not mention the materials' composition separately, thereby making the extraction challenging. For example, some of the materials have their compositions indicated within the IDs in an abbreviated form~\cite{NAZABAL200173} and did not mention them explicitly (see Fig. \ref{fig:id2comp}(a)). We also found tables where the composition of the materials is not specified; instead, their standard names are used as IDs. Such examples include Wollastonite or Diopside~\cite{schneider200029si}, which have a fixed chemical composition that can be obtained from standard sources/databases. In some cases, the composition was specified separately, but the IE model failed to extract the composition correctly due to dependency on material IDs, as shown in \ref{fig:id2comp}(b). Here, the variable `M' needs to be substituted by elements like `W', `Nb', or `Pb',  which needs to be inferred using the material IDs mentioned in the first column of the illustrated table~\cite{KAUR2015153}.


\textbf{e. Variables used to represent compounds:} When a composition is expressed with variables such as $(70-x)$TeO$_2$+15B$_2$O$_3$+15P$_2$O$_5$+$x$Li$_2$O, where $x$ = 5, 10, 15, 20, 25 and 30 mol\%~\cite{KHAFAGY20081460}, it mostly denotes the variation of different compounds. However, in some articles, variables have been used to represent compound names instead of their values. One such example is RE$_{36}$Y$_{20}$Al$_{24}$Co$_{20}$ where RE = Ce, Pr, Nd, Sm, Gd, Tb, Er, Sc~\cite{ZHAO20091001}. This scenario is observed in 1\% of the tables, where DiSCoMaT~\cite{gupta-etal-2023-discomat} fails to extract the material compositions. Note that this particular case can be solved using GPT-4, but as DiSCoMaT performs better in composition extraction from tables than GPT-4~\cite{mascqa-2024}, and a pipeline of GPT-4 and DiSCoMaT is not feasible, hence, this still remains an open challenge.

\subsubsection{Extracting compositions from text}
Now, we discuss the challenges in extracting the compositions reported in the text of the MatSci research papers. We report our statistical findings based on the frequency of each challenge. We also use GPT-4 to extract the compositions from text. The prompts given to GPT-4 for composition extraction are provided in Table ~\ref{tab:gpt4}. Specifically, we have used gpt4-1106 model through the OpenAI Python library. The temperature was set to 0.0 for reproducibility.

\textbf{a. Different formats of compositions:} The compositions in materials literature do not adhere to a predetermined pattern and encompass several variations. This is in strict contrast to notations in chemistry, where IUPAC nomenclature is used. Some notable examples are as follows. 


\textbf{1.}  \textit{"Erbium-doped glasses with the molar composition 40GeO$_2$.10SiO$_2$.25Nb$_2$O$_5$.25K$_2$O, plus 0.1 to 4 mol\% of ErO1.5, were prepared using mixtures of the respective oxides (99.99\% purity), with exception of K$_2$O, which was added in the form of K$_2$CO$_3$"}~\cite{SANTOS20102677}.\\
\textbf{2.}  \textit{"Bulk samples of (Se$_{80}$Te$_{20}$)$_{100-x}$Ag$_x$ (0 $\leq$ x $\leq$ 4) system were prepared by conventional melt quenching technique. High-purity (99.999\%) elements with appropriate atomic percentages were sealed in a quartz ampoule (length $\sim$ 100 mm and internal diameter $\sim$ 6 mm) in a vacuum of 10 - 5 mbar"}~\cite{SINGH20122826}.\\
\textbf{3.}  \textit{"The samples having chemical composition of 2(Ca,Sr,Ba)O–TiO$_2$–2SiO$_2$ were examined. CaO, SrO, and BaO contents in the samples were varied as shown in Table 1. RO\% shows the molar percentage of CaO, SrO or BaO in total RO of CaO+SrO+BaO"}~\cite{TSUZUKU200250}.

\textbf{b. Extracting variable values in text:}  Extracting values from variables is challenging since the variable values are specified in different formats, with some present only in the text. For instance, consider the following sentence from a peer-reviewed manuscript. ~\cite{S002230931200734X} \textit{A series of tellurite glasses with nominal composition (80-x)TeO$_2$–xGeO$_2$–10Nb$_2$O$_5$–10K$_2$O, where x = 0, 10, 20, 30, 40, 50, 60, 70 and 80 mol\%, hereafter named 8T0G, 7T1G, 6T2G, 5T3G, 4T4G, 3T5G, 2T6G, 1T7G and 0T8G, respectively, were doped with 0.2 to 4 mol\% $ErO_{1.5}$}. 

Although GPT-4 understands the doping element, since the entire information is not present in the same sentence and the exact values of doping content are not specified, it does not extract the composition successfully.

Here, the $x$ values representing the compositions and the respective variable names are present only in the text. Appendix~\ref{app_622_comp_text}(c) shows a few instances of other composition formats with variables. However, it may be noted that if full information is present in the sentences, GPT-4 is able to extract information correctly for the cases where the compositions are given in the form of variables.

\textbf{c. Low recall in extracting compositions expressed with variables:} 28\% of the articles have compositions written with variables, of which 28.57\% does not provide any values for the variables in the text. Among the 71.53\% where values are present, 40\% of them do not mention the step size for the range of values taken by the variable. For example, consider the text representing a set of compositions as follows from a manuscript: \textit{x(0.75AgI:0.25AgCl):(1-x)(Ag$_2$O:WO$_3$), where 0.1$\leq$x$\leq$1 in molar weight fraction.}~\cite{S0167273804001882} The step size of 0.1 is mentioned nowhere in the text but could be inferred from the composition table present in the paper. Therefore, extracting only from the text in such cases leads to more errors, and this can be resolved by connecting the variables to the correct composition table containing the variable. GPT-4 takes the endpoints for substituting the values in the compositions. However, due to a lack of information, it does not extract complete compositions due to the lack of values between the extreme values.

\textbf{d. Recognition of full forms and abbreviations:} 
Instead of providing precise composition values, full forms are employed instead of abbreviations. Consider the following example.

\textit{"Lithium disilicate glass was prepared in 30 g quantity by heating stoichiometric homogeneous mixtures of lithium carbonate (99.0\%), Synth, and silica (99.9999\%), Santa Rosa, for 4 h at 1500 °C in a platinum crucible."}~\cite{SCHRODER2014163}. This text indirectly mentions the glass's composition as lithium disilicate without clearly mentioning the percentages or numbers. GPT-4 is able to infer the chemical formulas from chemical names but cannot infer the exact composition and its percentages from the sentence.


\textbf{e. Unstable and irrelevant composition extraction:} Unstable reagents and other irrelevant compositions which does not refer to the material are also identified as compositions due to a lack of robust parsers. $AlO_4$ is an unstable entity referring to the aluminum tetrahedral structure, while $SiO_2$ can be a composition. These undesired extractions can lead to a huge drop in the precision of the IE model, and separating them from the material composition is not easy. Only a domain expert, with the help of the source article, can confirm whether the extraction is relevant or not. 
GPT4 fails to differentiate compositions from unstable compounds. \\

It is worth noting that although GPT-4 can address some of these challenges, especially extraction from text, its closed nature makes it challenging to use it at scale and for custom applications. Some of the reasons are:\\
\textbf{1.} Often, the research documents could be highly sensitive, preventing their sharing with commercial models such as GPT-4.\\
\textbf{2. } The inability of GPT-4 to be combined with smaller predictive models like DiSCoMaT prevents exploiting excellent domain-specific models that extract information very accurately.\\
\textbf{3. }
The commercial nature of such models can make it prohibitive due to the expenses associated with the usage due to the large number of sentences to be analyzed in the research papers and any additional prompt-engineering involved.

Therefore, GPT-4 may not be an ideal baseline for IE at large scale from research publications.

\subsubsection{Extracting compositions from table and text jointly}

Extracting information from PI tables is more challenging than extracting from CI tables, as the incomplete information in the table regarding the composition should be inferred from the text. A detailed analysis of 50 PI tables revealed that 36\% of the tables have unique challenges and are not ''regular''. To clarify this point further, we discuss some of these challenges below while also defining a ``regular'' MCC-PI table in Fig~\ref{fig:mcc-pi}. We have cross-checked all the reported challenges in this section by using the DiSCoMaT\cite{gupta-etal-2023-discomat}, the best IE model for composition extraction from MatSci tables~\cite{mascqa-2024}, which also handles PI tables; and found that the model was unsuccessful in extracting composition from tables having these characteristics.

\textbf{a. Unusual variables used:} Other than the common variables like $x$, $X$, $y$, $z$, and $Z$, we also encounter variables like $R$, $A$, $Y$ and $S$ in 4\% of the manuscripts. Distinguishing some of them, such as S or Y, is difficult as they are valid symbols for chemical elements as well~\cite{HENRY20031}. 

\begin{figure}
    \centering
    \includegraphics[width=0.9\textwidth]{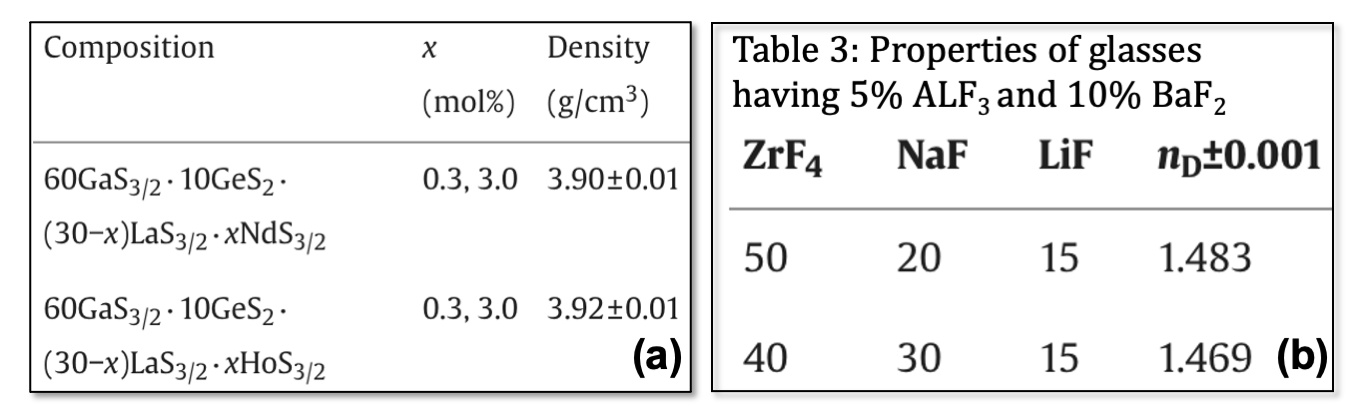}
    \caption{(a) Composition across multiple columns. (b) Partial composition in the table, rest in the text.}
    \label{fig:2_comp_new}
    \vspace{-0.2in}
\end{figure}


\textbf{b. Composition present across multiple columns}: The composition of the material is spread across multiple columns/rows (instance depicted in  Fig.~\ref{fig:2_comp_new}(a)~\cite{KADONO199939}), or the table does not follow any fixed orientation. This is observed among 4\% of the PI tables. 

\textbf{c. Composition partly in the table and partly in text:} Although PI tables contain the composition partly, it is expected that the complete information is available in the text. But in rare occurrences, as depicted by Fig. \ref{fig:2_comp_new}(b), we observe that only the remaining part of the composition, which is not mentioned in the table, is present in the text. This makes linking the parts of compositions in the text and tables challenging. Thus, extracting the whole composition is extremely difficult, a case seen in less than 1\% of the PI tables~\cite{le1997alkali}.

\textbf{d. Presence of multiple variables:} We found 6\% of the PI tables having more than one variable, all of which need to be taken into account to extract the composition correctly. As discussed previously, variables can be of various forms, making extracting multiple variables a challenging task~\cite{saddeek2010structural, hayashi1998thermal}.
\subsection{Extracting properties from tables}
\label{prop_table_extraction}
Until now, we focused on the extraction of compositions from tables and text. In this section, we discuss the challenges with property extraction. To this extent, we analyzed 100 arbitrarily selected property tables. The observations based on this analysis are as follows.

\begin{figure}
    \centering
    \includegraphics[width=0.95\textwidth]{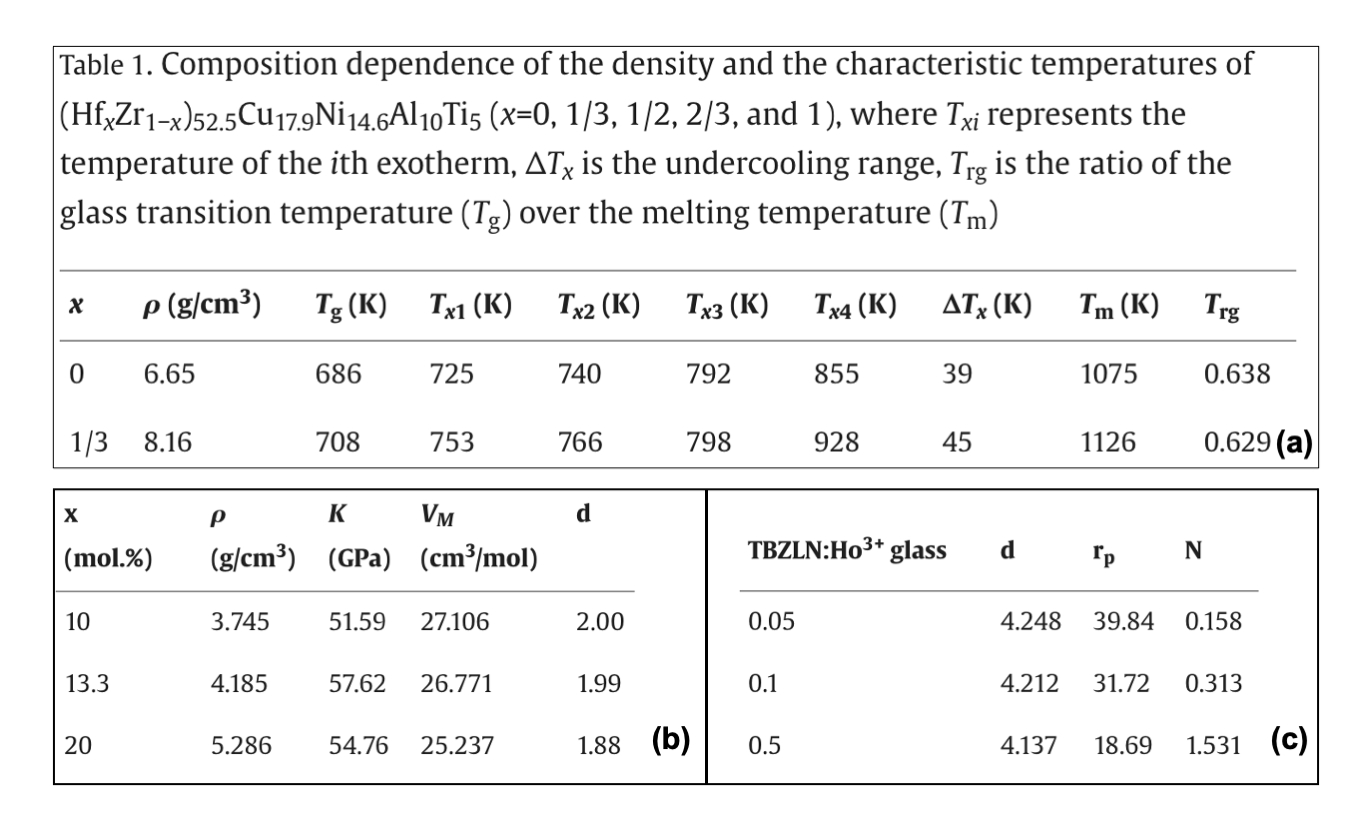}
    \caption{(a) Property description in caption \& semantically close headings, (b) Variable \textbf{`d'} representing fractal bond connectivity, (c) Variable \textbf{`d'} representing density.}
    \label{fig:Prop_caption}
\end{figure}

\textbf{a. Semantically similar row/column headers:} 19\% of the tables have similar abbreviations or headers with similar descriptions for different properties. For example, in Fig.~\ref{fig:Prop_caption}(a), the headings of the columns are T$_g$, T$_{x1}$, T$_{x2}$, T$_{x3}$, $\Delta${T$_{x}$}, T$_{m}$ ~\cite{GU200277}. Identifying the desired property by a predictor model or someone without domain knowledge can be difficult in this case. 

\textbf{b. The same property measured under different conditions:} The same property can be measured with different techniques or under different conditions. Therefore, it is important to extract the correct contextual information related to the reported property. Some recurrent scenarios include witnessing tables with various refractive indexes (RIs) at different wavelengths~\cite{QIAO2005357} (see Fig. \ref{fig:all_challenges}), glass transition temperatures at different heating rates~\cite{GIRIDHAR1998225}, or hardness at different testing loads. We encountered 9\% of the property tables exhibiting this challenge.

\textbf{c. Information in caption/footer instead of tables:} Often, properties are mentioned with abbreviations in the headings of tables, which are semantically close to other properties (for example, Fig.~\ref{fig:Prop_caption}(a)). The information regarding their abbreviation is commonly found in the caption or footer of the table. We observed 30\% of the tables having this characteristic~\cite{SCUDINO2005856,DEPARIS20052166}. Further, 2\% of the tables have no information on the properties units. However, these are found in the caption or footer of the tables~\cite{OKUNO20051032}. Hence, text from these sections might be handy for extracting our desired properties.



\textbf{d. Property recorded under various acronyms:} It is a common practice to record property names with their abbreviations. Some properties can have various abbreviations like density is represented with either $\rho$ or d, Young's modulus with YM or E, and activation energy with E$_0$, Ae, or E$_a$.

\textbf{e. Identical acronyms representing different entities:} We encountered tables (see Fig.~\ref{fig:Prop_caption}(b,c)) where the commonly used acronyms are used to represent different entities; not the usual property they generally represent. For example, `n', which is mostly used to represent RI, is also used to represent equation parameters specific to the experiments. Another commonly seen instance is `d' which is used to represent density~\cite{seshadri2014study} and has also been used to represent fractal bond connectivity~\cite{abd2014correlation}, lattice parameters, and equation parameters. This suggests that using a string-matching IE algorithm can result in poor performance in such cases.

\textbf{f. Range of values (min-max) given instead of mean value:} In very few cases (< 1\% tables), we encountered property values reported in range rather than a single value. For example, the values of T$_g$ are reported in the range 930-945$^{\circ}$C~~\cite{BESSON2000187}. Only a domain expert would know which value to take for a corresponding property between the min, max, or mean of the documented values. This might depend on the property or the application intended to be used, and will also be reflected in the IE algorithm.





\subsection{Challenges common for both composition and property extraction:}
\label{common_comp_prop_table_extraction}
\vspace{-0.1in}
Thus far, we discussed the challenges faced during composition extraction in~\ref{comp_table_extraction} and property extraction in~\ref{prop_table_extraction} from tables. However, some challenges arise in either of these scenarios.

\begin{figure}
    \centering
    \includegraphics[width=0.9\textwidth]{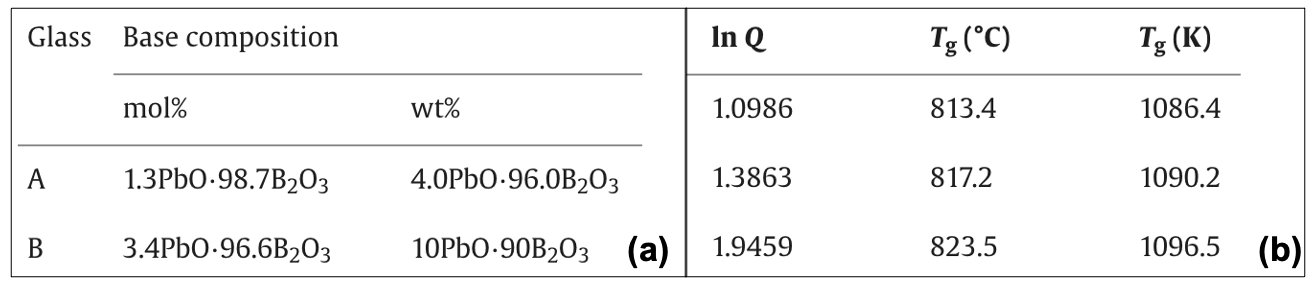}
    \caption{(a) The same glass composition mentioned in both mol\% and wt\%. (b) The same property of a material is mentioned multiple times with different units.}
    \label{fig:diff_units_comp}
\end{figure}

\textbf{a. Same composition or property represented with different units:} Tables are sometimes (2\%) presented with the essential information recorded in multiple units in different columns/rows. Fig.~\ref{fig:diff_units_comp} shows a composition table having composition in both mol\% and wt\%~\cite{KONISHI200019}, and a property table having glass transition temperature (T$_g$) mentioned in both $^{\circ}$C and K~\cite{ROCHERULLE199851}. This can lead to duplication of the extracted data.

\textbf{b. Multiple ways of reporting the same unit:} Despite the well-known and accepted conventions for writing the SI units~\cite{nist_guide}, research publications resort to multiple ways of reporting the same unit. For instance, for g/cm$^3$, several variations are observed in peer-reviewed publications such as gm/cm$^3$, g.cm$^{-3}$, g/cm$^3$, gcm$^{-3}$, g/cc, gm/cc, gw/cm$^3$, gm cc$^{-1}$. Similar observations are made for kg/m$^3$, where variations such as kgm$^{-3}$, kg/m$^3$, kg m$^{-3}$ are presented. Extracting the correct unit and normalizing it to a standard form is an essential task. Thus, while there are standard rules for writing SI units, it is observed that these are not strictly followed in scientific publications.

\begin{figure}
    \centering
    \includegraphics[width=0.60\textwidth]{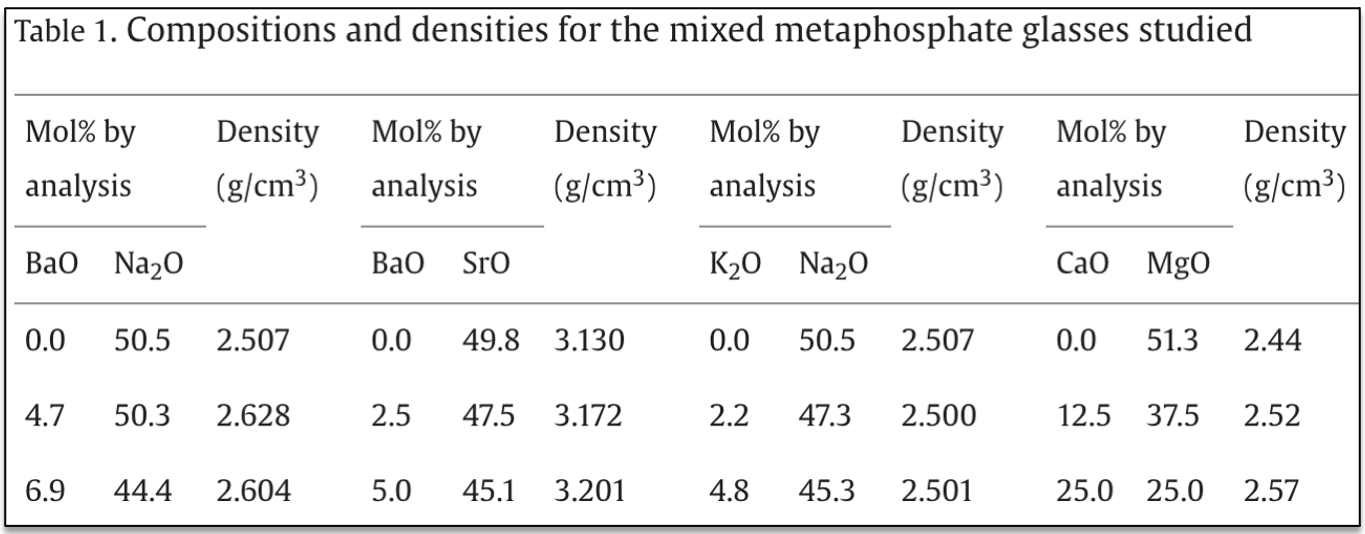}
    \caption{Multiple tables concatenated to form a larger table.}
    \label{fig:multi_tables}
\end{figure}

\textbf{c. Multiple tables merged in one:} A rarely seen challenge (<1\%) is illustrated in Fig.~\ref{fig:multi_tables}, where many tables are concatenated in a long or broad table, which leads to difficulties in extracting the required details~\cite{WALTER200048}. 

Note that none of these challenges could be solved using the IE model DiSCoMaT~\cite{gupta-etal-2023-discomat} and GPT-4.

\vspace{0.1in}
\subsection{IE for manufacturing and characterizing materials}

To identify the challenges in extracting precursors, processing and testing conditions, and material structure, we analyzed 50 arbitrarily selected papers from the dataset for reporting our findings.

\textbf{a. Precursor extraction:} A research paper generally investigates materials of a similar kind. Hence, it has to be assumed that all the materials are manufactured using the same precursors. In research papers where batch composition is mentioned in tables, the challenges are similar as mentioned in Section~\ref{comp_table_extraction}(b). In papers where researchers discuss the patented materials, they refer to them by their trademark name, for example, \textbf{\textit{Pyrex, BOROSIL, Gorilla,}} etc., and hence their precursor information is not provided. However, papers discussing materials reported in previous publications, provide references to those papers reporting the required information in detail.

\begin{figure}
    \centering
    \includegraphics[width=0.60\textwidth]{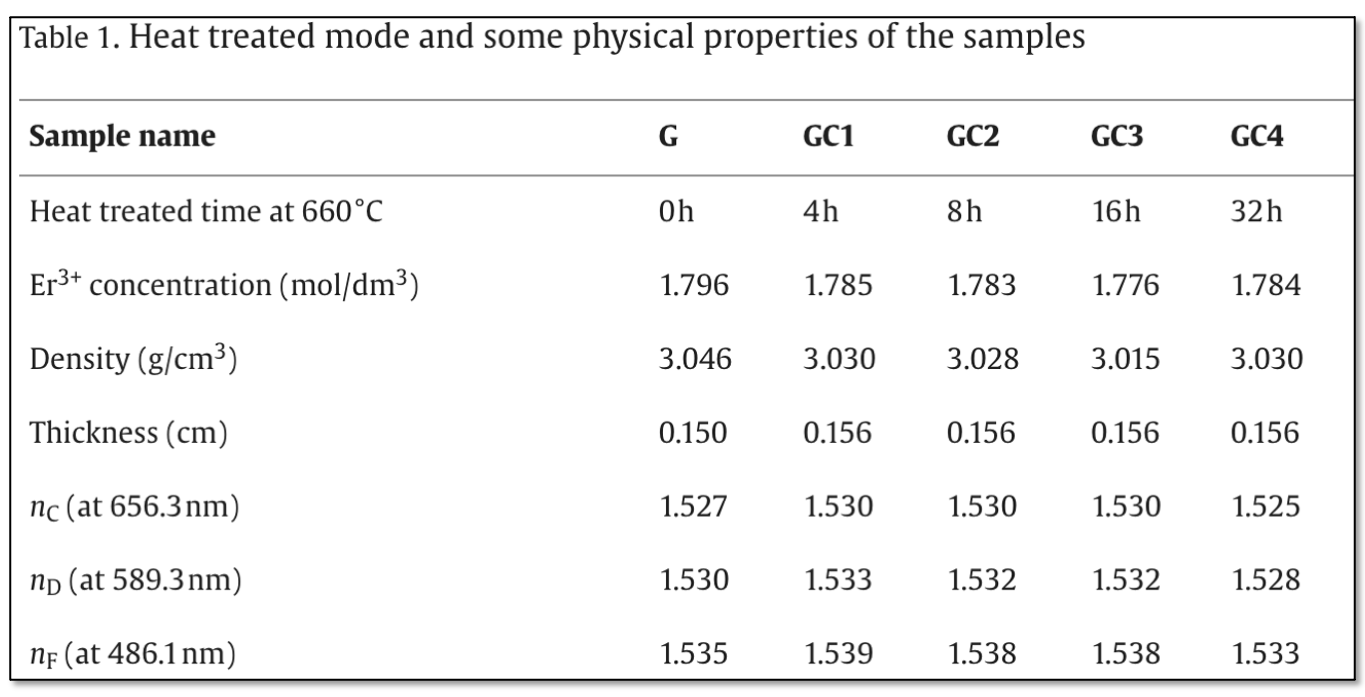}
    \caption{Challenges related to extraction of processing condition (heat treatment time) and property (refractive index) reported under various testing conditions (wavelength).}
    \label{fig:all_challenges}
    \vspace{-0.2in}
\end{figure}

\textbf{b. Processing conditions extraction:} Processing conditions reporting could be extremely non-linear and convoluted. Consider the set of sentences describing the processing conditions \cite{processing} as follows. \textit{``… powders were weighed and mixed thoroughly before being transferred to a \textbf{90 Pt/10 Rh crucible}, \textbf{heated} at \textbf{320°C} and \textbf{maintained} between \textbf{1000 and 1400°C} depending on composition, for approximately \textbf{25 min}. After \textbf{annealing} for approximately \textbf{three hours}, the glass was allowed to \textbf{cool slowly to room temperature}…''}. Hence, the challenges here are to extract temperatures and duration for each process, like heating, annealing, and cooling, along with the environmental conditions and experimental apparatus. Sometimes, these conditions are also mentioned in the table (see Fig. \ref{fig:all_challenges}), and their extraction poses similar challenges as described in Section \ref{prop_table_extraction}(b).

\textbf{c. Testing conditions extraction:} The testing conditions mainly comprise the sample characteristics, dimensions, test name, instrument name, instrument settings, and testing variables like temperature, wavelength, load, frequency, pressure, etc. Consider the following excerpt from \cite{sigoli2001phase}: \textit{``The porous microstructure of the matrix was investigated by scanning electron microscopy \textbf{(SEM) (JEOL JSM T330A)}, by \textbf{infrared spectroscopy (IR)} in a \textbf{FT-IR spectrometer (Perkin Elmer Spectrum 2000)}, and by \textbf{X-ray powder diffractometry (XRD) (Siemens D-5000)}. The phase separation process was investigated by Raman microscope. The \textbf{room temperature Raman measurements} were performed through \textbf{Raman imaging microscope (Renishaw) system 3000}, with the \textbf{632.8 nm He–Ne laser} line for excitation''}. The boldface text indicates the information to be extracted for obtaining a complete understanding of the testing process of a material. Fig.~\ref{fig:all_challenges} lists different wavelengths at which a material is tested to obtain refractive index. The challenges faced in IE for this case will be similar to the ones posed in Section \ref{prop_table_extraction}(b).



\textbf{Material structure:} To study the structure of materials, researchers perform X-ray diffraction studies, obtain the Raman spectra, optical micrographs, and scanning electron micrographs depending upon the depth of detail about the material structure required. This information is mostly reported in figures and the figure description in the text provides some important details about the material's structure. In the statement, ``\textit{The \textbf{Raman spectrum} of the \textbf{porous} phase (Fig. 6(b)) shows only one \textbf{band at 277 cm$^{-1}$} assigned to \textbf{silica} vibrations...}''\cite{sigoli2001phase}, the information about Raman spectra is already shown in the graph, and the text mentions only critical findings.

To summarise, the extraction of precursors, processing, and testing conditions from text poses challenges related to named entity recognition and relation extraction, which requires the need for specialized datasets and model development. there exist several materials science domain-specific models capable of extracting this information but their performance (F1-Score) on different types of desired entities ranges from as low as 33\% \cite{gupta2022matscibert} (interlayer materials for batteries, taken from SOFC dataset\cite{friedrich-etal-2020-sofc}) to 93\% \cite{matbert_trewartha2022quantifying}(materials tag, taken from MatScholar dataset\cite{matscholar_weston2019named}). There also exist some knowledge graphs created using these tools like MatKG\cite{venugopal2022matkg}, however, the quality of the information in such sources is as good as the underlying model. Further, on relation-extraction tasks, the best-performing models have an F1 score of 0.82\cite{song-etal-2023-matsci}, which indicates significant efforts required to facilitate the information extraction and complete materials science tetrahedron. Further, the extracted entities should be linked with the respective materials. The challenges faced during IE from tables for processing and testing variables require overcoming similar challenges as explained earlier for composition (Section \ref{comp_table_extraction}) and properties (Section \ref{prop_table_extraction}).

\subsection{MatSci Knowledge-base: Linking extracted information}
The tetrahedron, as shown in Fig. \ref{fig:graph_abs} will be considered complete for a given material if its properties, processing, testing conditions, and raw materials required to manufacture are available.  To this end, researchers need to link extracted compositions with these variables. These pose unique challenges as it requires linking information among different entities within the paper such as connecting different paragraphs of the paper, text with tables, or tables with other tables in the paper.

\begin{figure}
    \centering
    \includegraphics[width=0.90\textwidth]{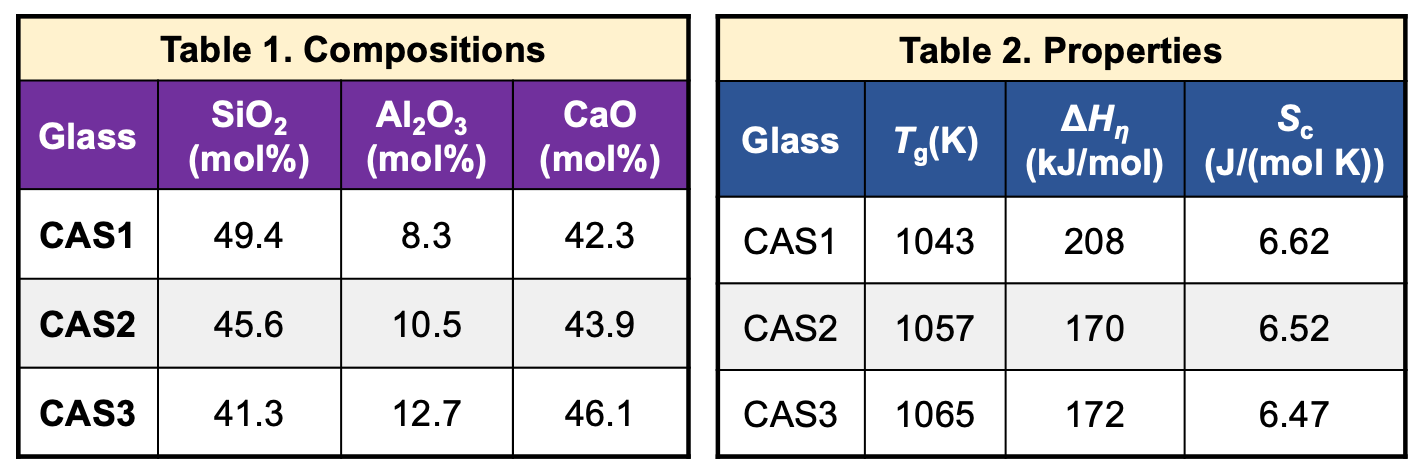}
    \caption{Composition and properties of the same material are mentioned in different tables within the article.}
    \label{fig:inter_table_kausik}
\end{figure}

Material IDs are required to link information across multiple tables. For instance, in Fig. \ref{fig:inter_table_kausik}~\cite{solvang2004rheological}, we obtain the composition of CAS1 from Table 1 and $T_g$ of this material from another table(Fig. \ref{fig:inter_table_kausik} Table 2.). Every material in an article should have a unique ID, which should be used consistently across the whole article to denote the corresponding material. Any exception to this will lead to difficulties in linking our extracted information. We detected 187 out of 2536 (7.37\%) publications where inter-table IE is necessary and found difficulties in 81 of them while connecting the different components of the tetrahedron.

\textbf{a. Different material IDs in different tables:} The same materials have been reported with different IDs in different tables. 21 out of the 81 research papers (25.93\%) have this challenge~\cite{ROMERO2000106,MARTINELLI2000263}.\\
\textbf{b. Material IDs absent from tables:} We detected MatSci papers where no IDs are present in the tables. There exist 23 out of 81 (28.40\%) documents having this challenge, where compositions of the materials and their corresponding properties are reported in separate tables, but neither of the tables have any ID present denoting the material. ~\cite{PEITL199939}.\\
\textbf{c. One of the tables does not contain material IDs:} While connecting two tables, there are cases where IDs are mentioned only in one table ~\cite{BEGGIORA2003476} (37 out of 81 (45.67\%) papers with this challenge).

As we observe that material ID is a very important factor in connecting tables, we did an intensive analysis of the type of IDs that are reported in the tables (see Table.~\ref{tab:id_table}).

\vspace{0.1in}
\subsection{ID Analysis}

\begin{figure}
    \centering
    \includegraphics[width=0.65\textwidth, page=11]{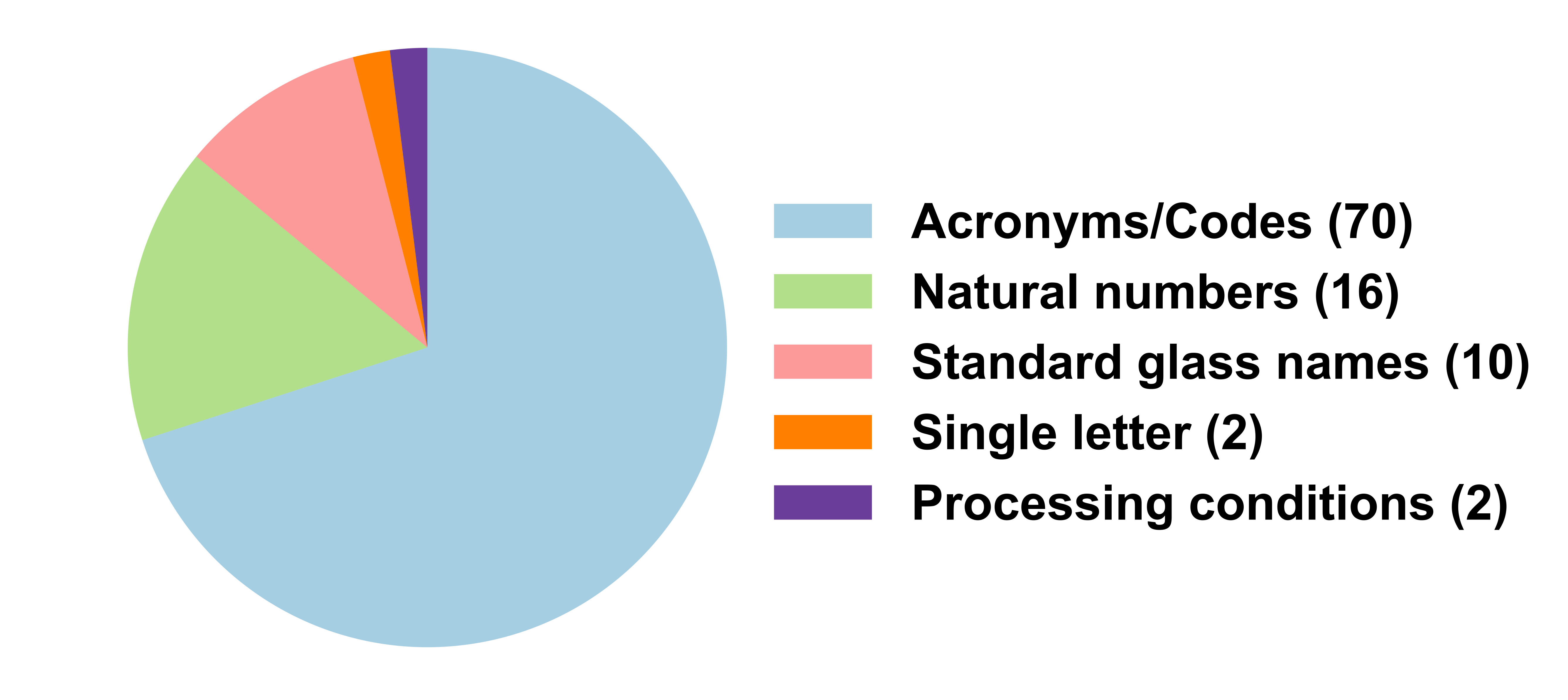}
    \caption{Writing styles of IDs in MatSci articles.}
    \label{fig:pie_chart}
\end{figure}

As material ID is the key component in connecting materials from tables to text, across two different tables, or also across different sections of the text, we investigated arbitrarily selected 50 articles containing material IDs in the tables and recorded their semantic pattern to observe the semantics used by authors to refer to materials. We found that a majority of the authors prefer to use acronyms or self-made codes as IDs for referring to the materials, followed by natural numbers and standard material names, illustrated in Fig.~\ref{fig:pie_chart}. Material IDs are generally present at the beginning of the table and very rarely seen in the middle or end. Often, we come across tables having IDs that contain relevant information like the processing conditions of the material, or information about the state of the glass like amorphous or crystalline, or its composition, which are not separately mentioned in the table. These information are generally encoded as abbreviations, and extracting them can be challenging. In Table~\ref{tab:id_table}, we describe different cases along with the percentage occurrences. Note that the composition of the material should not be confused with its ID, as both are separate entities. An ID is expected to be shorter in length, most likely an acronym, and unique to each material.

\begin{table}[!htb]
\centering
\scalebox{0.9}{
\begin{tabular}{cc}
\toprule
\textbf{Challenges in IE from material IDs} & \textbf{\% of occurence} \\
\midrule
Composition info/doping conc. present only in IDs & 20 \\
IDs present in the middle & 2 \\
Multiple IDs present for the same composition & 4 \\
State or structural info in ID & 2 \\
Info or references about the processing conditions & 8 \\
Same IDs but different composition & 4 \\
The article contains IDs interconnected & 2 \\
Taken from other articles & 6 \\
\bottomrule
\end{tabular}
}
\vspace{0.1in}
\caption{Different challenges in extracting information from material IDs and their occurrences.}
\label{tab:id_table}
\end{table}




\section{Guidelines for writing IE-friendly MatSci tables}

Tables should be reported in such a way that automated extraction and the detection of the desired information are easy. Some of our suggestions are as follows (illustrated with Fig.~\ref{fig:perfect_table}, adapted from \cite{kosuge1998thermal}):

\begin{figure}
    \centering
    \includegraphics[width=0.90\textwidth]{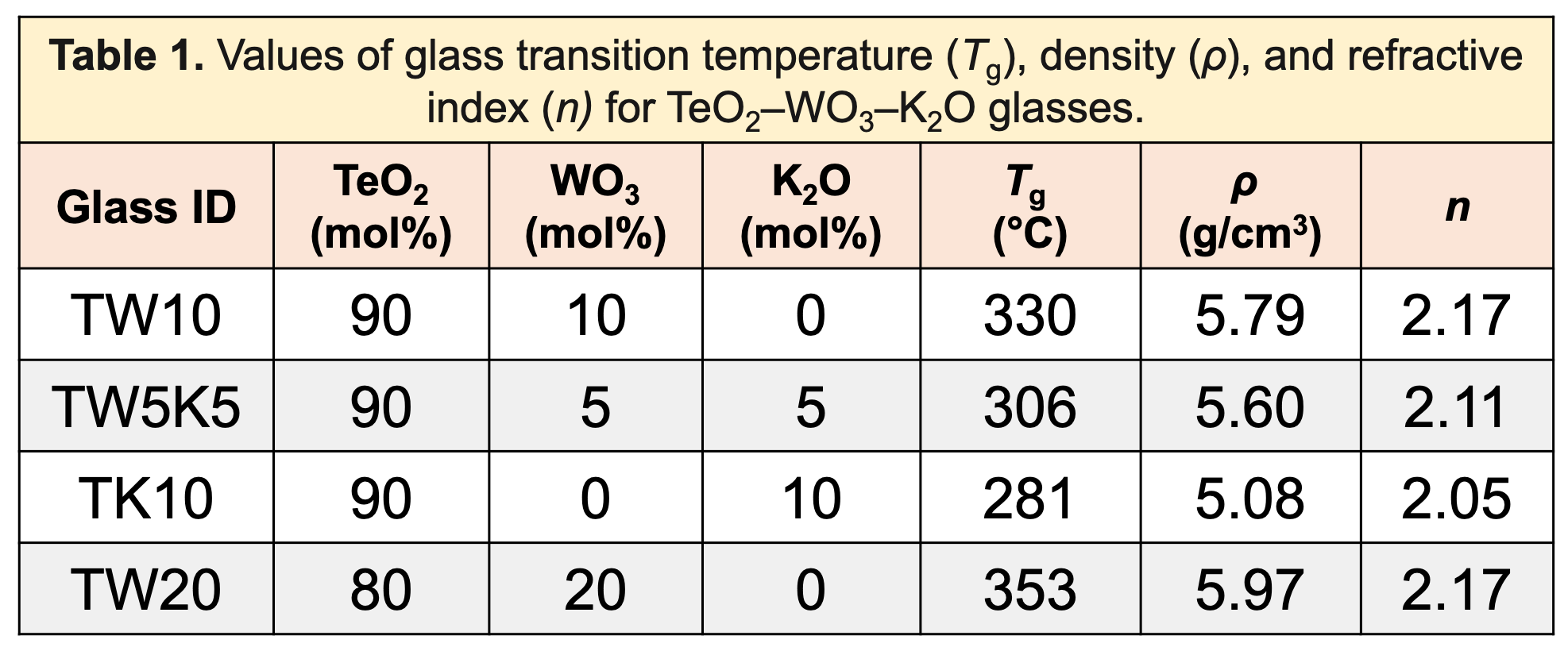}
    \caption{IE-friendly MatSci table.}
    \label{fig:perfect_table}
\end{figure}

\textbf{a. Use column orientation:} Many IE algorithms that have been developed for tables have considered column orientation only. Also, we showed that 93\% of the published tables are column-oriented. The following suggestions assume that we are following column orientation. \\
\textbf{b. Use MCC-CI tables:} Tables should have the components associated with a composition written in different cells. Moreover, the table should have the complete information of the material compositions (see Fig~\ref{fig:class_comp_tab}). \\
\textbf{c. Use proper and descriptive headers:} The headers should contain the chemical formula of the compounds or elements that make up the materials, along with the acronyms of the reported properties, with processing, and testing conditions. If precursors, processing, and testing conditions are common, they can be omitted from tables.\\
\textbf{d. Use standard notations for units:} Units should be mentioned in the column headers of the tables within brackets. Moreover, the standard notations for representing the SI unit should be consistently used.\\
\textbf{e. All-in-one table:} Prefer writing all the information of a particular material in a single table while following proper orientation. Following this will avoid the need for inter-table extraction. \\
\textbf{f. IDs are mandatory:} Material IDs are important to identify different materials mentioned in the tables and link them across tables and text. IDs should be mandatory for tables and written in the first column.\\
\textbf{g. Consistent IDs:} Material IDs should be formed as an acronym of its comprising constituents. They should be consistent in the whole article, that is, there should not be more than one ID referring to the same material.\\
\textbf{h. Table structure:} Table should be of the structure [[Material ID], [C1], [C2], ..., [P1], [P2], ...]. `C' denotes the constituting compound/elements that form the material. Sequence them so that their proportions are arranged in a descending manner. `P' refers to the properties of the corresponding material.\\
\textbf{i. Column/row-wise consistency:} Each column or row should contain information only related to a particular entity mentioned in the heading or in the first row, respectively. Multiple tables or columns should not be concatenated into one.\\
\textbf{j. Captions:} All tables should have a clear, concise, and descriptive caption. The table caption should clearly explain the acronyms used in the tables.





\section{Conclusion and future work}
\vspace{-0.1in}
The literature is replete with IE challenges and algorithms to extract information about materials. However, there exists no study that quantifies how much benefit can be obtained if a particular challenge is solved. In this paper, we have identified and quantified several unresolved challenges present in IE for every aspect of the MatSci tetrahedron. Specifically, we pointed out the locations in a MatSci research paper where each piece of information on the MatSci tetrahedron of a given material is reported. Further, we outlined the challenges associated with IE and linking them to build the MatSci KB. We hope this extensive analysis will motivate researchers to focus on the challenges in the field, giving an insight into the gain associated with each of these challenges. This will also enable the researchers to identify the right problems to focus on based on the desired outcome. Finally, we provided recommendations for an IE-friendly table format to enhance the automated extraction of the desired information and improve the researchers' tabular understanding. Such concerted efforts are required to streamline the reporting in MatSci articles, thereby accelerating IE for materials discovery.

\section{Author Contributions}
\vspace{-0.1in}
Hira, K - Conceptualization, Methodology, Software, Validation, Investigation, Data Curation, Writing - Original Draft.\\
Zaki, M - Conceptualization, Methodology, Validation, Investigation, Visualization, Data Curation, Writing - Original Draft.\\
Sheth, D. - Investigation, Data Curation, Writing - Original Draft.\\
Mausam - Conceptualization, Writing - Review \& Editing, Supervision, Project administration.\\
Krishnan, N.M.A. - Conceptualization, Writing - Review \& Editing, Supervision, Project administration.\\


\section{Conflicts of interest}
\vspace{-0.1in}
There are no conflicts to declare.

\section{Acknowledgements}
N. M. Anoop Krishnan acknowledges the funding support received from BRNS YSRA (53/20/01/2021-BRNS), Google, and Intel Labs. Mohd Zaki acknowledges the funding received from the PMRF award by the Government of India. Mausam acknowledges grants by Google, IBM, Verisk, and a Jai Gupta chair fellowship. He also acknowledges travel support from Google and Yardi School of AI travel grants. The authors thank the High Performance Computing (HPC) facility at IIT Delhi for computational and storage resources. We thank the anonymous reviewers of the 2023 AI4Mat NeurIPS workshop for recommending our work to be published in the special issue of Digital Discovery.

\newpage

\bibliography{main}
\bibliographystyle{vancouver}

\newpage

\section{Appendix}


In this section, we will address some more notable challenges, most of which have been solved satisfactorily by IE models.\\
The details of all the research papers used in this study, along with annotations to identify the challenges, are available at ~\url{https://github.com/M3RG-IITD/MatSci-IE-Challanges}

\subsection{Common Challenges faced during information extraction from tables}

We begin by discussing the problems encountered for all-encompassing IE tasks. Challenge \textbf{\textit{a}} has been resolved in ~\cite{gupta-etal-2023-discomat} while challenge \textbf{\textit{b}} has been addressed by ~\cite{habibi2020deeptable}~\cite{gupta-etal-2023-discomat}.

\textbf{a. Distractor rows or columns:} Additional contents in the table that are irrelevant to our desired information.

\textbf{b. Different orientations of tables:} Each table can have either of the two orientations - row or column, which is essential to recognize for extracting information precisely. We saw 100 random composition tables and 100 random property tables and observed that 7\% of the tables are represented with row orientation (see Fig.\ref{fig:all_challenges}), whereas 93\% of the tables are represented with column orientation (see Fig. \ref{fig:mcc-pi}).



\subsection{Other challenges faced in composition extraction:}

\subsubsection{From tables:}

\begin{wrapfigure}{r}{0.50\textwidth}
    \vspace{-0.4in}
    \centering
    \includegraphics[width=0.50\textwidth]{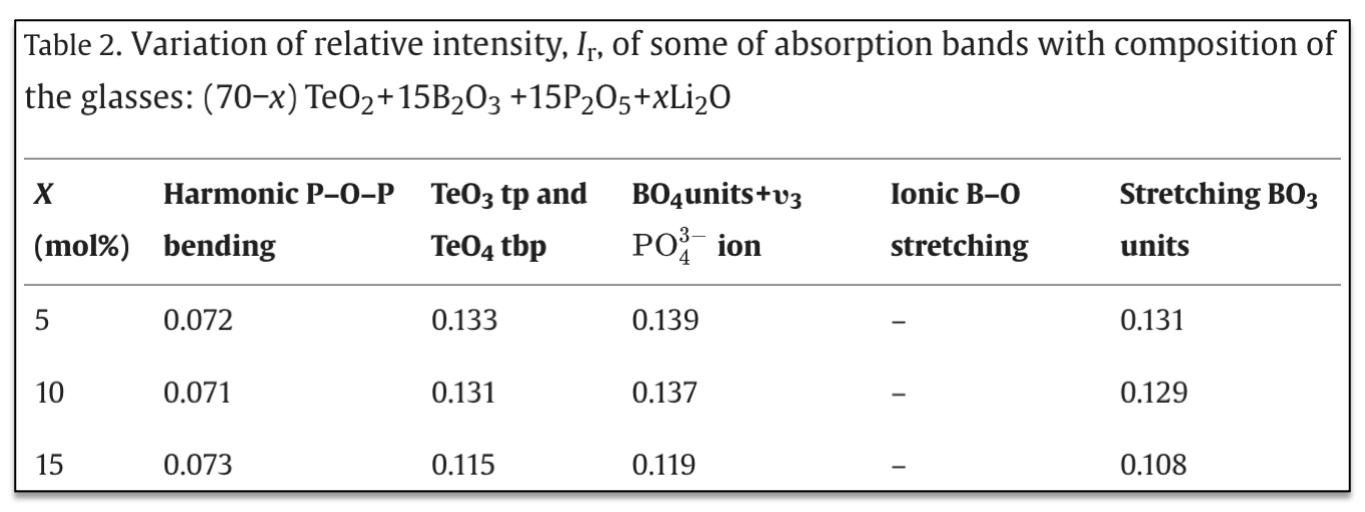}
    \caption{MCC-PI table with variable `x'}
    \label{fig:mcc-pi}
    \vspace{-0.2in}
\end{wrapfigure}

We start by illustrating a typical MCC-PI~\cite{KHAFAGY20081460} in Figure \ref{fig:mcc-pi} table without any challenges for the reader's convenience.


\vspace{.05in}

We discuss three more challenges which can be seen in the composition tables. Challenge \textbf{a} and \textbf{b} has already been handled by Gupta et al. ~\cite{gupta-etal-2023-discomat}. In challenge \textbf{c}, extraction of compositions mentioned with atomic\%, atomic fraction and parts per million (ppm) is still outstanding, whereas extraction of dopant concentration from challenge \textbf{d} has not been solved yet.

\textbf{a. One Composition with multiple units:} Consider the following example composition - 0.85TeO$_2$+0.15WO$_3$+0.1wt\%Ag$_2$O+0.076wt\%CeO$_2$ ~\cite{DUCLERE20092195}. Here, for a given material, different components are measured in different units (mol\% and wt\%). This is found in 2\% of the tables making composition extraction challenging.

\textbf{b. Composition in table headers:} Most tabular IE models like Tabbie~\cite{iida2021tabbie}, DiSCoMaT~\cite{gupta-etal-2023-discomat} perform better when row/column headers contain appropriate information regarding its contents. In MatSci tables, the headers are mostly material IDs, compound names, properties, processing and testing labels, and the inner cells contain corresponding values. However, in 6\% of the tables, we found that the compounds with their values were present in the heading, which makes it hard for the IE models to extract the desired information. For instance, Se$_{58}$Ge$_{33}$Pb$_9$~\cite{DEEPIKA20091274} or $x$ = 10\%, $x$ = 20\%,...~\cite{MINATI2007502} are column headers which contain both the compounds and corresponding concentration in the heading. 67\% of these were SCC-CI, whereas the rest 33\% were MCC-PI tables.  

\textbf{c. Composition expressed with different units in various articles:} such as mol\%, weight\%, atomic\%,  mol fraction, weight fraction, atomic fraction, and ppm. Among them, the most commonly used unit is mol\%, followed by weight\%.

\textbf{d. Percentage not equal to 100}: In some papers, even after extracting the whole composition correctly, we observe that the sum of the chemical component concentrations is not equal to 100, whereas we also notice the presence of the scenario where composition is extracted incorrectly and the sum is equal to 100. Especially in the case of doping, the sum exceeds 100, which is correct. The challenge is to identify where we need to normalize the values extracted and where we should not. We noted that dopant is reported in 2\% of the composition tables.




\subsubsection{From text}
\label{app_622_comp_text}
Both \textbf{a} and \textbf{b} are unsolved. In challenge \textbf{b}, we do not know whether the extracted composition needs to be normalized or it is partially extracted. Normalization is not a challenge after correct extraction as there are existing works on it\cite{gupta-etal-2023-discomat}, but currently, no work has been done on extracting the composition completely if it is not fully mentioned in the text.

\textbf{a. Unit not mentioned:}  39.53\% compositions had no unit specified explicitly. 

\textbf{b. Percentages not summing to 100:} Out of 78\% compositions found in the text, 17.94\% of them did not have the sum of values of the chemical compounds equal to 100. 

\textbf{c. Different formats of compositions with variables:} A few instances of different formats of compositions expressed in variables are:\\
\textbf{1.} The non-isothermal crystallization kinetics of xLi$_2$S–(1-x)Sb$_2$S$_3$, x=0–0.17 were investigated using differential scanning calorimetry (DSC).~\cite{DELAPARRA2003104} \\
\textbf{2.} To ascertain the effect of the glass composition on fluorescence parameters around 1.86 $\mu$m, we prepared and experimented on two series of glasses. The first one was aR$^1$$_2$O(1-a)TeO$_2$ where 'a' was 0, 10, 15, 20, 30 mol\%, and 'R$^1$' was Li, Na, K. The second one was bR$^{11}$O.cR$_2$$^{111}$O$_3$(1-b-c)TeO$_2$ where 'b' was 0, 10, 20, 30 mol\%, and 'c' was 0.5\% or 16.5\%, and 'R$^{11}$' = Ba, 'R$^{111}$' = Al, Ga, or In. To find the effect of concentration quenching, the concentration of thulium oxide was varied from 0.01 to 5.0 mol\%~\cite{HOLLIS2001422}. \\
\textbf{3.} Glasses with composition in mol\%: 51ZrF$_4$, 16BaF$_2$, 5LaF$_3$, 3AlF$_3$, 20LiF, 5PbF$_2$ have been prepared by melting of the powders (commercial raw materials of purity higher than 99.99\%) in a covered vitreous carbon crucible at about 850 °C for 45 min in a dry argon glove box with a water content lower than 5 ppm. The melt was poured into a preheated copper mould at 240 °C and slowly cooled down to room temperature. The doping ion was added in excess to the formula +xErF$_3$ from 0.01 to 11 mol\% corresponding to 0.02 to 22 × 1020 Er3+ ions/cm3. The samples obtained were of good optical quality~\cite{MORTIER2003505}.


\subsection{From table and text jointly:}

\begin{wrapfigure}{r}{0.50\textwidth}
    \vspace{-0.2in}
    \centering
    \includegraphics[width=0.50\textwidth]{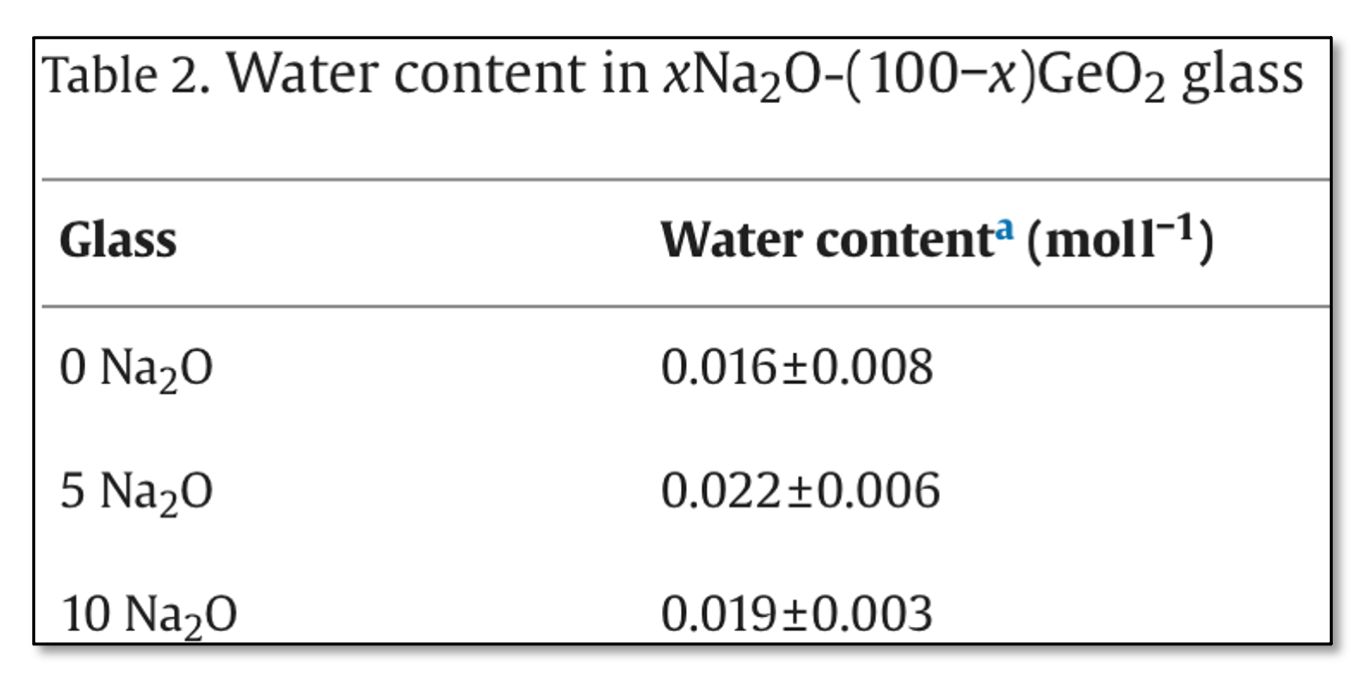}
    \caption{Variable \textbf{`x'} is not in table. **}
    \label{fig:2_comp}
    \vspace{-0.1in}
\end{wrapfigure}
\vspace{.05in}

\textbf{a. Variables representing composition in text not found in tables:} A generic way of extracting the composition when an arithmetic equation of the composition containing variables is mentioned in the text is to connect it to the variables located in the headings of the table and substitute it with values mentioned under it. Extraction becomes difficult if there is an absence or mismatch between the variable name in the table and the text, as shown in Fig.~\ref{fig:2_comp}. We found 8\% of the tables posing this challenge~\cite{YOSHIDA200328}. This challenge has been resolved in the IE model proposed by Gupta et. al.~\cite{gupta-etal-2023-discomat}. 

** Please note that we have mistakenly added Fig. \ref{fig:mcc-pi} as Fig.~\ref{fig:2_comp} in our Digital Discovery Publication, although we have cited the correct source in the text. We sincerely apologise for this mistake.

\begin{longtable}{|p{3cm}|p{7cm}|p{3cm}|}
\caption{GPT-4 prompt and responses for composition extraction from text}\\
\hline
\textbf{Prompt} & \textbf{GPT-4 Response} & \textbf{Conclusion} \\
\hline
\endfirsthead
\hline
\textbf{Prompt} & \textbf{GPT-4 Response} & \textbf{Conclusion} \\
\hline
\endhead
“Extract all the compositions from the following expression. (70 - x)TeO2 + 15B2O3 + 15P2O5 + xLi2O, where x = 5, 10, 15, 20, 25 and 30 mol\%”  & To extract all the compositions based on the given expression for different values of x (5, 10, 15, 20, 25, and 30 mol\%), substitute each value of x into the expression and calculate the resulting composition. 

The expression given is:  
(70 - x) TeO2 + 15B2O3 + 15P2O5 + xLi2O 

Let's substitute the values of x one by one and write down the resulting compositions. 
  
For x = 5 mol\%: 
(70 - 5)TeO2 + 15B2O3 + 15P2O5 + 5Li2O = 65TeO2 + 15B2O3 + 15P2O5 + 5Li2O 
  
For x = 10 mol\%: 
(70 - 10)TeO2 + 15B2O3 + 15P2O5 + 10Li2O = 60TeO2 + 15B2O3 + 15P2O5 + 10Li2O 
  
For x = 15 mol\%: 
(70 - 15)TeO2 + 15B2O3 + 15P2O5 + 15Li2O = 55TeO2 + 15B2O3 + 15P2O5 + 15Li2O 
  
For x = 20 mol\%: 
(70 - 20)TeO2 + 15B2O3 + 15P2O5 + 20Li2O = 50TeO2 + 15B2O3 + 15P2O5 + 20Li2O   

For x = 25 mol\%: 
(70 - 25)TeO2 + 15B2O3 + 15P2O5 + 25Li2O = 45TeO2 + 15B2O3 + 15P2O5 + 25Li2O 
  
For x = 30 mol\%: 
(70 - 30)TeO2 + 15B2O3 + 15P2O5 + 30Li2O = 40TeO2 + 15B2O3 + 15P2O5 + 30Li2O & GPT-4 is able to extract information correctly for the cases where the compositions are given in the form of variables. \\
\hline
“Extract all the compositions from the following expression. Glasses with composition in mol\%: 51ZrF4, 16BaF2, 5LaF3, 3AlF3, 20LiF, 5PbF2 have been prepared. +xErF3 from 0.01 to 11 mol\%” & The given expression lists the components of a glass system in terms of their mole percentage (mol\%). Additionally, it mentions that the glass composition can be doped with varying amounts of ErF3 (erbium fluoride) from 0.01 to 11 mol\%. The base composition of the glasses is as follows:

- ZrF4 (zirconium fluoride): 51 mol\%
- BaF2 (barium fluoride): 16 mol\%
- LaF3 (lanthanum fluoride): 5 mol\%
- AlF3 (aluminium fluoride): 3 mol\%
- LiF (lithium fluoride): 20 mol\%
- PbF2 (lead(II) fluoride): 5 mol\%

Additionally, we have the ErF3 dopant, which ranges from 0.01 to 11 mol\%.

A specific glass composition would include a certain percentage of ErF3 within the given range, and because the total mol\% must sum up to 100\%, adding ErF3 will decrease the percentages of the other components proportionally.

Here's how to calculate the adjusted percentages for a specific amount of ErF3 added:

Let's say we add x mol\% ErF3 to the glass:

- The sum of the other components without ErF3 is 100\%
- After adding x mol\% ErF3, the sum of the original components must be scaled down to (100 - x)\%
- Each of the original components will be scaled by multiplying their percentages by (100 - x)\% / 100

For example, for a composition with 2 mol\% ErF3:

- Total of the original components: 100\% - 2\% ErF3 = 98\%
- ZrF4: 51 mol\% * (98/100) = 49.98 mol\%
- BaF2: 16 mol\% * (98/100) = 15.68 mol\%
- LaF3: 5 mol\% * (98/100) = 4.90 mol\%
- AlF3: 3 mol\% * (98/100) = 2.94 mol\%
- LiF: 20 mol\% * (98/100) = 19.60 mol\%
- PbF2: 5 mol\% * (98/100) = 4.90 mol\%

So the new composition with 2 mol\% ErF3 added would be approximately:

- ZrF4: 49.98 mol\%
- BaF2: 15.68 mol\%
- LaF3: 4.90 mol\%
- AlF3: 2.94 mol\%
- LiF: 19.60 mol\%
- PbF2: 4.90 mol\%
- ErF3: 2 mol\%

For every specific addition of ErF3 within the range of 0.01 to 11 mol\%, this calculation would have to be performed to determine the adjusted composition. The exact values would of course depend on the precise amount of ErF3 being added to the composition. & GPT-4 understands the doping element, but since the entire information is not present in the same sentence, as well as exact values of doping content is not specified, it is able to give just the partial information. \\
\hline
“Extract all the compositions from the following expression. Lithium disilicate glass was prepared in 30 g quantity by heating stoichiometric homogeneous mixtures of lithium carbonate (99.0\%), Synth, and silica (99.9999\%), Santa Rosa, for 4 h at 1500 °C in a platinum crucible." & The compositions mentioned in the given expression include:

1. Lithium Carbonate (Li2CO3) with a purity of 99.0\% - This is one of the starting materials used to prepare the lithium disilicate glass.

2. Silica (SiO2) with a purity of 99.9999\% - This is the other starting material used in the preparation. "Santa Rosa" is likely a brand name or location associated with the high-purity silica used.

There are no exact quantities provided for each individual component (lithium carbonate and silica); however, the overall quantity of the prepared lithium disilicate glass is given as 30 g. The process involves heating these stoichiometric mixtures at 1500 °C for 4 hours in a platinum crucible, which avoids contamination as platinum is inert to the process. 

To determine the stoichiometry, you would need the molecular weights of the reactants and the targeted lithium disilicate (Li2Si2O5) product. Even though exact proportions are not given in this expression, you can infer that the molar ratios are configured to produce lithium disilicate without excessive amounts of either lithium carbonate or silica. "Synth" could represent that the lithium carbonate used is synthetic or refers to a brand or grade, but no additional context for this term is provided. & GPT-4 was able to understand the chemical compounds but was unable to produce exact composition percentages for the same. \\
\hline

“Extract all the compositions from the following expression. The formation of SiO3 groups was also confirmed by electron diffraction analysis on thin (4–80 Å) films of pure SiO2 .The strong cohesive energy of the SiO4 groups (previously noted during studies of displacement cascades in complex glasses [21]) makes them less ‘vulnerable’ than BO4 or AlO4 groups, which more easily give up an oxygen atom." & From the provided expression, the compositions mentioned are:

1. SiO3 groups
2. SiO2 (pure silicon dioxide)
3. SiO4 groups
4. BO4 groups
5. AlO4 groups

These represent various oxide groups or arrangements of silicon (Si), boron (B), and aluminum (Al) with oxygen (O). & GPT4 fails to differentiate compositions from the unstable compounds. \\
\hline
\label{tab:gpt4}
\end{longtable}

\end{document}